\documentclass[nohyperref]{article}

\usepackage{microtype}
\usepackage{graphicx}
\usepackage{subfigure}
\usepackage{booktabs} 

\usepackage{hyperref}

\usepackage[accepted]{icml2022}

\usepackage{amsmath}
\usepackage{amssymb}
\usepackage{mathtools}
\usepackage{amsthm}

\usepackage[capitalize,noabbrev]{cleveref}

\theoremstyle{plain}

\theoremstyle{definition}

\theoremstyle{remark}

\usepackage{enumitem}

\usepackage[textsize=tiny]{todonotes}

\usepackage{makecell}
\usepackage{multirow}
\usepackage{colortbl,hhline}
\definecolor{light-gray}{gray}{0.95}
\usepackage{wrapfig,lipsum,booktabs}

\begin{document}

\twocolumn[
\icmltitle{Meta Knowledge Distillation}




\begin{icmlauthorlist}
\icmlauthor{Jihao Liu}{sensetime}
\icmlauthor{Boxiao Liu}{sensetime}
\icmlauthor{Hongsheng Li}{cuhk}
\icmlauthor{Yu Liu}{sensetime}
\end{icmlauthorlist}

\icmlaffiliation{cuhk}{The Chinese University of Hong Kong \& Centre for Perceptual and Interactive Intelligence}
\icmlaffiliation{sensetime}{SenseTime Research}

\icmlcorrespondingauthor{Hongsheng Li}{hsli@ee.cuhk.edu.hk}


\vskip 0.3in
]



\printAffiliationsAndNotice{}  

\begin{abstract}
Recent studies pointed out that knowledge distillation (KD) suffers from two degradation problems, the teacher-student gap and the incompatibility with strong data augmentations, making it not applicable to training state-of-the-art models, which are trained with advanced augmentations. However, we observe that a key factor, i.e., the temperatures in the softmax functions for generating probabilities of both the teacher and student models, was mostly overlooked in previous methods. With properly tuned temperatures, such degradation problems of KD can be much mitigated. However, instead of relying on a naive grid search, which shows poor transferability, we propose Meta Knowledge Distillation (MKD) to meta-learn the distillation with learnable meta temperature parameters. The meta parameters are adaptively adjusted during training according to the gradients of the learning objective. We validate that MKD is robust to different dataset scales, different teacher/student architectures, and different types of data augmentation.
With MKD, we achieve the best performance with popular ViT architectures among compared methods that use only ImageNet-1K as training data, ranging from tiny to large models. 
With ViT-L, we achieve 86.5\% with 600 epochs of training, 0.6\% better than MAE that trains for 1,650 epochs. 
\end{abstract}

\section{Introduction}
\label{introduction}

Knowledge distillation (KD) \cite{kd} is an extensively studied technique in recent years, whose initial goal is to to transfer the ``dark knowledge" from a teacher model to a student model. It works by providing the student with more intuitive targets than the ground truth one-hot labels to learn. It has been widely explored in various fields, including model compression \cite{beyer2021knowledge}, semi-supervise learning \cite{mpl}, domain adaptation \cite{crd}. Besides of those applications of KD, many other works aims at improving the vanilla KD via introducing extra supervisions, including intermediate features \cite{fitnet}, attention maps \cite{zagoruyko2016paying}, relations \cite{park2019relational}, etc. 


However, recent papers pointed out that knowledge distillation suffers from two problems: (1) \textbf{Teacher-student gap} \cite{takd, earlystopkd} Even though the performance of a teacher model can improve, the performance of the student model does not necessarily follow the same trend. (2) \textbf{The incompatibility with strong augmentation} \cite{das2020empirical,cui2021isotonic} When the teacher's performance improves via strong augmentation, the performance of the distilled student model might actually degrade. To demonstrate the two problems, we conduct a series of pilot studies. We train ResNet56 teacher models with different data augmentations, and use ResNet20 to distill those teacher models, with results shown in Table \ref{tab:augmentation}. Besides, we train a series of ResNet models with different depths, and use those teachers to distill ResNet20 student, with results shown in Table \ref{tab:teacher}. In Table \ref{tab:augmentation}, we can see that the strongest teacher (trained with CutMix \cite{yun2019cutmix}) actually results in worse student performance than the teacher trained with normal data augmentation. Besides, as shown in Table \ref{tab:teacher}, the distilled student's performance decline with the improvement of teacher's performance.

\begin{table}[t]
    \centering
    \caption{\textbf{Knowledge distillation is not compatible with strong augmentation.} \cite{das2020empirical,cui2021isotonic}}
    \resizebox{0.9\linewidth}{!}{
    \begin{tabular}{lc|c}
        Augmentation & Teacher acc. & Student acc. \\
        \midrule
        Normal & 73.3 & \textbf{70.4} \\
        LabelSmooth \cite{inception} & 73.3 & 68.1 \\
        Mixup \cite{zhang2017mixup} & 74.3 & 67.9 \\
        CutMix \cite{yun2019cutmix} & \textbf{76.0} & 68.8 \\
    \end{tabular}
    }
    \label{tab:augmentation}
\end{table}

\begin{table}[t]
    \centering
    \caption{\textbf{Knowledge distillation suffers from teacher-student gap.} \cite{takd, earlystopkd}}
    \resizebox{0.9\linewidth}{!}{
    \begin{tabular}{lc|c}
        Teacher & Teacher acc. & Student (ResNet20) acc. \\
        \midrule
        ResNet32 \cite{resnet} & 71.0 & \textbf{70.8} \\
        ResNet56 \cite{resnet} & 73.3 & 70.4 \\
        ResNet110 \cite{resnet} & 75.4 & 70.5 \\
        ResNet164 \cite{resnet} & \textbf{75.8} & 69.9 \\
    \end{tabular}
    }
    \label{tab:teacher}
\end{table}


However, in the pilot studies, we observe that a key factor, the temperature hyperparameters, $\tau$, of both the teacher and student, were mostly overlooked by previous methods. It is actually the main cause for the two above mentioned problems. All the mentioned problems reported by previous methods are based on experiments with a fixed recommended temperature, i.e. 4 in \cite{crd, earlystopkd}. The temperature is used to control the degree of softness of the targets. 
With lower temperatures, the student pays much more attention to match the maximal logits of the teacher outputs. On the other hand, higher temperatures encourage the student to also focus on the logits other than the maximal ones. 
With different augmentations or teacher architectures, the distribution of the teachers' output logits may vary significantly. 
We empirical find the soften logits of a teacher trained with CutMix is too noisy to provide effective supervision for distillation.
This problems become severe for larger teacher models, because those models are usually trained with augmentations of data deformations of larger degrees. 
Besides, prior arts \cite{kd} assigned the same temperature to both the student and the teacher, trying to match them with the same degree of softness. These approaches do not take the capacity gap between the teacher and the student into consideration. For example, it might be harder for a student (e.g. ResNet18 \cite{resnet}), to distill from a larger teacher (e.g. ResNet101) than a smaller one (e.g. ResNet50), given that the same temperature is used for both the teacher and student.


One of our key findings is that, with a proper temperature, the aforementioned degradation problems in KD can be much mitigated. We show that vanilla KD can beat state-of-the-art distillation methods on distilling vision transformers \cite{vit}.

Previous works \cite{beyer2021knowledge, earlystopkd} simply conducts a grid search to identify an optimal temperature, and fix it for the entire learning process.
In this paper, we propose Meta Knowledge Distillation (MKD) to learn the distillation function with learnable meta-parameters $\phi$, including the temperatures $\{{\tau}_t, {\tau}_s\}$ of the teacher and the student, respectively.
The meta-parameters are adjusted online during the students' training process with a meta objective that minimizes the validation loss on a preserved validation set, allowing the distillation function to dynamically adapt over time to the gradients of learning objective.
We apply MKD to various distillation scenarios, and obtain consistent gains over the manually provided meta-parameters suggested by previous methods. Empirically, we validate that MKD is robust to different scales of the datasets, architectures of the teacher or student models, and different types of data augmentation. While previous work \cite{deit} shows that vanilla KD does not work on Vision Transformer (ViT), we show that MKD improves ViT consistently. We achieve the best accuracy of ViT without changing architectures when training on ImageNet-1K \cite{imagenet} dataset only. 
Specifically, we achieve 77.1\% top-1 accuracy with ViT-T, 2\% better than Manifold Distillation \cite{manifold}. We also achieve 86.5\% top-1 accuracy with ViT-L, surpassing the reported performance 85.15\% in \cite{vit}, which is trained with 10x larger ImageNet-21k \cite{imagenet}.

In summary, our contributions are as follows:
\begin{enumerate}
    \item We identify the temperature hyperparameters of both the teacher and student are overlooked by previous studies of the problems of knowledge distillation, and find that proper temperatures can mitigate those problems.
    \item We propose MKD to learn the distillation function by dynamically adapting the temperatures over time to the gradients of the training objective.
    \item With MKD, we achieve the best accuracy of ViT without changing architectures when training on ImageNet-1K dataset only, surpassing the results trained with 10x larger dataset. 
\end{enumerate}

\begin{figure*}[t]
    \centering
    \begin{minipage}[b]{0.68\linewidth}
        \centering
        \includegraphics[width=0.95\linewidth]{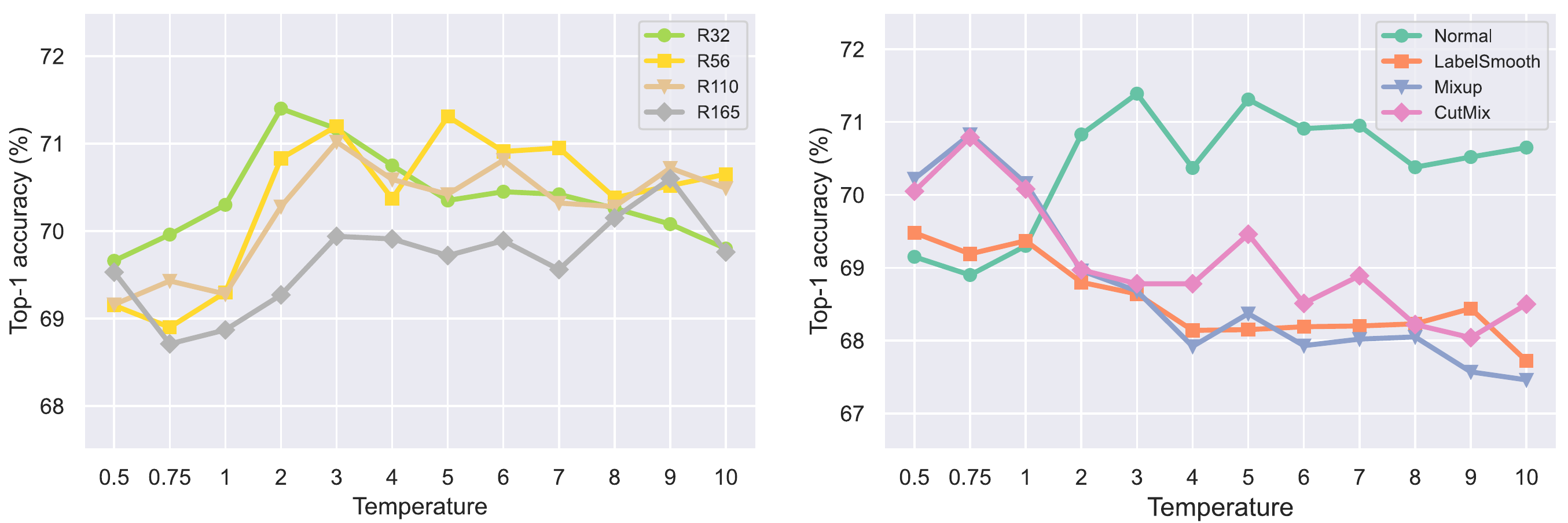}
        %
        \caption{\textbf{Grid search over temperatures of different teachers.} On the left, the results shown with different lines are distilled from the teachers trained with different data augmentations. On the right, different lines represent ResNet teachers with different depths.}
        \label{fig:grid_search}
    \end{minipage}
    \hspace{0.1em}
    \centering
    \begin{minipage}[b]{0.25\linewidth}
        \centering
        \includegraphics[width=1.0\linewidth]{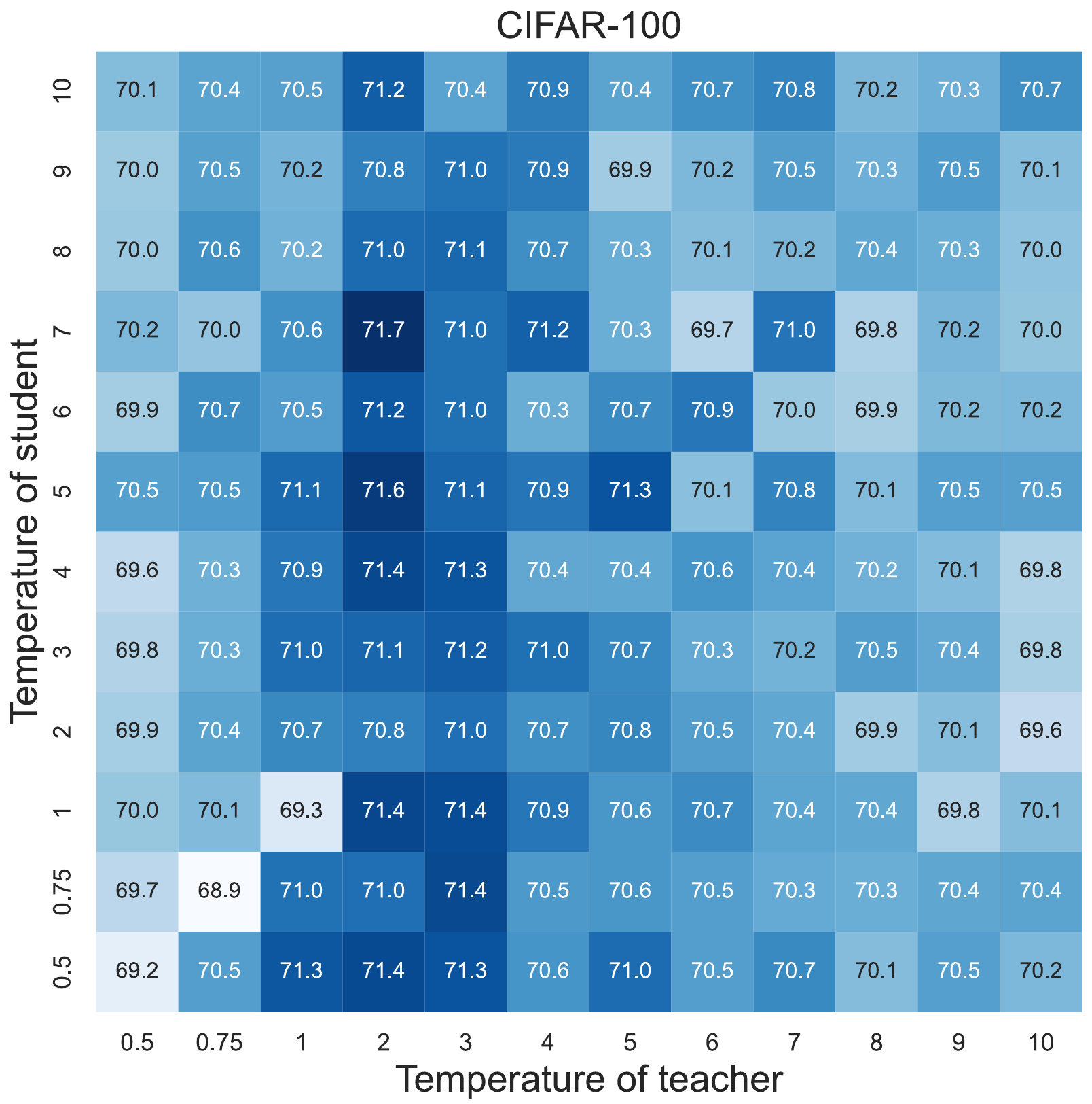}
        \vspace{-2em}
        \caption{\textbf{Grid search results of applying different temperatures to the teacher and the student.}}
        \label{fig:confusion}
    \end{minipage}
\end{figure*}


\section{Temperature Matters for Knowledge Distillation}
\label{sec:temperature}

Although the problems of knowledge distillation have been widely discussed in different literatures (as stated in Section \ref{introduction}), we find that the conclusions were reached by previous methods based on a fixed temperature. A properly tuned temperature can sometimes lead to opposite conclusions. We perform a grid search over temperatures to verify their impacts on the performance of KD. We train ResNet56 teacher models with different data augmentations, and use ResNet20 to distill those teacher models. Besides, we train a series of ResNet models with different depths, and use those teachers to distill ResNet20 student. The setting is the same as Tables \ref{tab:augmentation} and \ref{tab:teacher}, but different temperatures are applied. We use the same temperature for the teacher and the student by default.

As shown in Figure \ref{fig:grid_search}, the optimal temperatures vary for different teachers, and we can obtain opposite conclusions under different temperatures. 
For example, in Figure \ref{fig:grid_search} (left), R32 is the best teacher when the temperature is 4, but the worst teacher when using temperature of 9. Besides, in Figure \ref{fig:grid_search} (right), the teacher using normal data augmentation performs best when using temperatures of 3 among other teachers, but performs worst when the temperature is set as 0.75.

We also grid search different temperatures for both the teacher (ResNet56) and student (ResNet20), with results shown in Figure \ref{fig:confusion}. We find that using the same temperature for the teacher and the student leads to sub-optimal results, even sometimes the worst results. As shown in Figure \ref{fig:confusion}, the worst performance is obtained when using the same temperatures 0.75 for both the teacher and student.

We notice some recent papers try to study the compatibility of label smooth and KD. M{\"u}ller et al. \yrcite{muller2019does} showed that teachers trained with label smoothing lead to inferior student networks compared to teachers trained with hard targets. In contrast, Shen et al. \yrcite{shen2021label} reached a different conclusion showing that teachers trained with label smoothing produce better students. In our study, we can reach both conclusion by using different temperatures. As shown in Figure \ref{fig:grid_search} (left), when using lower temperatures (i.e. $\tau \leq 1$), the teacher trained with label smoothing produces slightly better results. However, when using higher temperatures (i.e. $\tau > 1$), the conclusion is opposite.

The temperatures for the teacher and the student play important roles in KD, and the optimal temperatures vary a lot under different distillation setups, as shown in our empirical studies. To address this problem, we introduce our Meta Knowledge Distillation in Section \ref{sec:mkd}.

\section{Meta Knowledge Distillation}
\label{sec:mkd}

In contrast to prior work improving knowledge distillation, which has sought to design new distillation loss \cite{crd}, new distillation pipeline \cite{takd}, and new distillation targets \cite{mpl}, we aim to improve knowledge distillation \cite{kd} via meta learning the temperature parameters to automatically adapt to various datasets, teacher-student pairs, and augmentations.


\subsection{Problem Formulation}

In the original formulation of knowledge distillation (KD) \cite{kd}, a student model is trained to match the targets provided by a teacher model, which tries to provide more information than the ground truth one-hot labels. 
The targets are usually softened by a fixed temperature in the softmax function and the entropy increases as the temperature increases.

Formally, we consider a student model $\mathcal{S}$ with parameters $\theta_s$ for distilling the teacher. We denote the output logits from the teacher and the student as $z_s$ and $z_t$, respectively. The student is trained to minimize the cross-entropy (CE) loss between its predicted probability $p_s$ and the teacher's output probability $p_t$, where $p_s$ and $p_t$ are softened by their corresponding temperatures ${\tau}_s$ and ${\tau}_t$ respectively.
\begin{align}
        & \min_{{\theta}_{s}} ~ \mathrm{CE}(p_s, p_t), \label{eq:obj}\\
        \text{where} ~~ p_s=~&\mathrm{softmax}(z_s/\tau_s), \ p_t=\mathrm{softmax}(z_t/\tau_t). \nonumber
\end{align}

In prior arts, $\tau_s$ and $\tau_t$ are manually chosen as the same constant greater than 1. However, as we discussed in Sec. \ref{sec:temperature}, the temperatures play a key role in distillation, and the optimal temperatures vary in different distillation setups, which are difficult to tune manually. Besides, the temperatures control the degree of softness of the targets provided by the teacher. Using a fixed temperature for different distillation setups, including teach/student architectures, augmentation types, datasets, can be sub-optimal.
Intuitively, the temperatures are expected to be adjusted according to according to the different training setups.

We take an explicit approach to tackle this problem: we aim at meta-learning the temperatures of both the teacher and student to dynamically adjust the softness of the learning targets according to the student's on-the-fly performance on the validation set,
\begin{equation}
    \label{eq:meta_obj}
    \begin{aligned}
    \min_{\phi} &\quad \mathcal{L}_v(\theta_s, \phi), \\
    \mathrm{s.t.} &\quad \theta_s = \underset{\theta_s}{\mathrm{argmin}} \,\, \mathcal{L}_t(\theta_s),
    \end{aligned}
\end{equation}
where $\mathcal{L}_t$ and $\mathcal{L}_v$ denote the training and validation losses, and $\phi = \{\tau_s, \tau_t\}$ are meta-parameters (temperatures) to be optimized on the meta level.

\subsection{Meta-learning Temperatures for Better Distillation}

Our Meta Knowledge Distillation (MKD) aims at jointly updating the student's parameters $\theta_s$ and temperature meta-parameters $\phi$. We apply the underlying distillation loss to a batch of training samples to pre-update the student. We then measure the performance of the pre-updated student with a batch of validation samples to update the meta-parameters. The student would be finally optimized with the updated meta-parameters. 

Specifically, starting from the current $\theta_s$ and $\phi$, MKD first pre-updates $\theta_s$ by optimizing the distillation objective using a batch of training samples, resulting in pre-updated parameters $\theta_s'$,
\begin{equation}
\label{eq:update}
    \theta_s' = \theta_s - \alpha\frac{\partial \mathcal{L}_t(\theta_s, \phi)}{\partial \theta_s},
\end{equation}
where $\alpha$ is the step size. 
Note that we approximate $\mathrm{argmin}_{\theta_s} \mathcal{L}_t(\theta_s)$ in Equation \eqref{eq:meta_obj} with an one-step update following Equation \ref{eq:update}, as updating $\theta_s$ until full convergence is infeasible.

MKD then measures the performance of the pre-updated parameters $\theta_s'$ with a new batch of validation samples, and utilizes the differentiable meta-objective $\mathcal{L}_v(\theta_s')$ to update $\phi$. 
The gradient of $\phi$ can be obtained by applying the chain rule of gradients,
\begin{equation}
\label{eq:gradient}
    \frac{\partial \mathcal{L}_v(\theta_s')}{\partial \phi} = \frac{\partial \mathcal{L}_v(\theta_s')}{\partial \theta_s'}\frac{\partial \theta_s'}{\partial \phi}.
\end{equation}

The meta-optimization is conducted via stochastic gradient descent, such that the meta-parameters $\phi$ are updated as
\begin{equation}
\label{eq:update2}
    \phi \leftarrow \phi - \beta\frac{\partial \mathcal{L}_v(\theta_s')}{\partial \phi},
\end{equation}
where $\beta$ is the meta step size. Intuitively, our proposed MKD aims at updating meta-parameters in the direction that minimizes the validation loss along the normal updates. 

After that, MKD updates the student's original parameters $\theta_s$ with the updated meta-parameters $\phi$,
\begin{equation}
\label{eq:update3}
    \theta_s \leftarrow \theta_s - \alpha\frac{\partial \mathcal{L}_t(\theta_s, \phi)}{\partial \theta_s}.
\end{equation}

In contrast to MPL \cite{mpl}, our approach does not reuse the student's update, as it make the student's updates follow those of the meta-parameters. We illustrate our method in Algorithm \ref{alg:mkd}.

\begin{algorithm}[tb]
   \caption{Meta Knowledge Distillaion}
   \label{alg:mkd}
\begin{algorithmic}[1]
    \REQUIRE student parametrized with $\theta_s$ and meta-parameters $\phi$
    \REQUIRE step size hyperparameters $\alpha$ and $\beta$
    \WHILE {not done}
        \STATE pre-update student: $\theta_s' = \theta_s - \alpha\frac{\partial \mathcal{L}_t(\theta_s, \phi)}{\partial \theta_s}$
        \STATE update meta-parameters: $\phi \leftarrow \phi - \beta\frac{\partial \mathcal{L}_v(\theta_s')}{\partial \phi}$
        \STATE update student: $\theta_s \leftarrow \theta_s - \alpha\frac{\partial \mathcal{L}_t(\theta_s, \phi)}{\partial \theta_s}$
    \ENDWHILE
\end{algorithmic}
\end{algorithm}

\subsection{Temperature Prediction Network}

Instead of directly updating the temperatures, we opt to make the tempuratures output by an extra small network, which takes a learnable embedding $e$ as input.
The extra small network is implemented as a 2-layer multi-layer perceptron (MLP) followed by a sigmoid function $\sigma$,

\begin{equation}
    \label{eq:generate_tau}
    \{{\tau}_s, {\tau}_t\} = \tau_{init} + \sigma(\mathrm{MLP}(e)) - 0.5,
\end{equation}
where $\tau_{init}$ denotes the initial temperature. 
We found that the extra small network has larger capacity than simply updating the two temperature and thus can adapt the training process much faster, leading to better performance.

\subsection{Alternative Meta-learning Objective}
\label{sec:meta_objective}

The meta-parameters $\phi$ are optimized to minimize 
the CE loss between the student's output and the ground-truth one-hot labels on a preserved validation set. However, in our experiments, we find that the CE loss might not truly indicate the actual performance. As long as the logit value of the correct class decreases, the loss decreases, but the performance does not necessarily increase. To address this problem, we also propose an alternative and more aggressive meta-objective that only optimizes the incorrect samples, 
\begin{equation}
\label{eq:meta_loss}
    \mathcal{L}_v(\theta_s) = \sum_{i\in\mathcal{G}}\sum_{j=1}^{c}{(p_{s}^{(ij)} - y^{(ij)})^2}, 
\end{equation}
where $\mathcal{G}$ denotes the indexes of the incorrectly classified samples in a validation batch, and $c$ denotes the number of classes, $p_{s}^{(ij)}$ returns the $j$th predicted probability value of $i$th sample, and $y^{(ij)}$ returns the $j$th ground-truth probability value of $i$th sample.

\section{Experimental Setup}
\label{sec:experimental_setup}
We apply Meta Knowledge Distillation (MKD) on various teacher-student pairs and augmentations setups, and compare MKD against state-of-the-art distillation methods on standard benchmarks such as CIFAR-100 \cite{cifar}, and ImageNet-1K \cite{imagenet}. 

\vspace{-3pt}
\paragraph{Datasets.} CIFAR-100 contains 50K images from 100 classes for training and 10K for testing. ImageNet-1K provides 1.2 million images of 1k classes for training and 50k images for validation. For those datasets, we preserve 10\% of the training set for optimization of the meta parameters, i.e. 5k images for CIFAR-100 and 128k for ImageNet-1K.

\vspace{-3pt}
\paragraph{Experimental setup.} 
For the small-scale CIFAR-100 benchmark, we follow CRD \cite{crd}, and experiment with different teacher-student pairs, including ResNet \cite{resnet} of different depths and widths. We mainly compare our results with vanilla KD \cite{kd} and state-of-the-art distillation method CRD \cite{crd}.

For large-scale ImageNet benchmark, we experiment with the state-of-the-art architecture, Vision Transformer (ViT) \cite{vit}, and mainly use the following three setups.
\begin{enumerate}
    \item For fair comparison with previous distillation methods for ViTs \cite{deit, manifold}, we employ the same teacher-student model pairs, i.e. ViT-T and ViT-S \cite{deit} to distill CaiT-S24 \cite{cait}.
    
    \item To compare with state-of-the-art ViT performances \cite{mae, trainvit}, we use a stronger teacher BEiT-L \cite{beit} with students ranging from ViT-T to ViT-L \cite{vit}. 
    \item We also experiment with a fixed student ViT-T, and use teachers of different performances (i.e. ViT-S, ViT-B, and ViT-L), to verify that whether a better teacher consistently leads to better distilled students.
\end{enumerate}

\vspace{-4pt}
\paragraph{Training details.} 

On CIFAR-100 benchmark, all the models are trained with SGD for 300 epochs, initial learning rate of 0.05, minimal learning rate of 0.0005 with cosine decay, weight decay of $5 \times 10^{-4}$. The teacher models are trained with CutMix \cite{yun2019cutmix}, while the students are not. Besides, we employ regular random crop and flip for data augmentation. For CRD, we use the nice official code \footnote{https://github.com/HobbitLong/RepDistiller} and the default hyperparameters of CRD. 

As ViT is sensitive to the choice of training hyper-parameters, we mainly follow the popular training recipe in \cite{deit}, but employ different dropping patch ratios \cite{droppath} for different students. We add a new distillation token to the student network for distillation. It works similarly to the class token \cite{vit} but its supervision is from the teacher. We do \textit{not} use extra network augmentations such as relative position \cite{beit} or layer scaling \cite{beit,cait}.
We distill ViT-T or ViT-S using a regular setting of training for 300 epochs, and distill ViT-base or ViT-large for 600 epochs (MAE \cite{mae} uses much more epochs, i.e. 1600). When comparing with previous best results, i.e. Table \ref{tab:sota}, we train ViT-S for 600 epochs.
In contrast to previous works that rely on a weight hyper-parameter to balance the cross-entropy loss term on ground-truth one-hot labels and the loss term on soft labels, we employ a simpler form that only minimizes the cross-entropy loss on soft labels from the teacher as Equation \eqref{eq:obj}. We employ AdamW optimizer \cite{adamw} to optimize meta-parameters, with a learning rate of $3 \times 10^{-4}$ and weight decay of $5 \times 10^{-4}$. For more details, please refer to the Appendix \ref{sec:appendix}.

For ViT, we find that the changing the temperatures during the training process is less important than the final converged temperatures, as ViT is a powerful network and can adapt to new temperatures quickly. Therefore, we fix and adopt the temperature learned from ViT-T. We only activate our proposed meta-learning strategy in the last 100 training epochs to maintain high training efficiency.

\section{Main Results}
\label{sec:main_results}

\subsection{Distilling ViT on ImageNet-1K}

\paragraph{Comparisons with ViT distillation methods.} In Table \ref{tab:compare_kd}, we compare Meta Knowledge Distillation (MKD)  with previous Vision Transformer (ViT) \cite{vit} distillation methods \cite{deit,manifold} using the same teacher-student pairs. MKD improves the performances by large margins (i.e. 4.2\% for ViT-T, and 2.3\% for ViT-S), indicating the potential of ViT when trained with better supervisions.

Previous work \cite{deit} shows that the vanilla KD does not work on ViT, and proposes a hard distillation method to pursue better performance. In contrast, our MKD is based on the vanilla KD, and achieves even better performance by adapting the temperatures. For instance, we achieve 76.4\% with ViT-T, +2\% and +4.2\% better than the hard and soft distillation strategies in \cite{deit}, respectively.

Manifold distillation \cite{manifold} designs more complex distillation targets, and requires specific teacher architectures. Comparing with their approach, our MKD is more accurate while being simpler and more flexible. Our MKD is able to be applied to various teacher-student pairs, and can take advantage of much stronger teacher, as shown in Table \ref{tab:fix_student}.

\begin{table}[t]
    \centering
    \caption{\textbf{Comparison with previous ViT distillation methods on ImageNet-1K.} All models are distilled with CaiT-S24 (82.7\%) as the teacher and trained from scratch with ImageNet-1K.}
    \resizebox{1.0\linewidth}{!}{
        \begin{tabular}{l|c|c}
        \toprule
        Distillation Method & Student & Top-1 Acc. (\%) \\
        \midrule
        No distillation & \multirow{5}{*}{ViT-T} & \underline{72.2} \\
        Soft (KD) \cite{deit} &  & 72.2 \\
        Hard \cite{deit} &  & 74.4 \\
        Manifold \cite{manifold} & & 75.1 \\
        \textbf{MKD} &  & \textbf{76.4} \\
        \midrule
        No distillation & \multirow{5}{*}{ViT-S} & \underline{79.8} \\
        Soft (KD) \cite{deit} &  & 80.0 \\
        Hard \cite{deit} &  & 81.3 \\
        Manifold \cite{manifold} & & 81.5 \\
        \textbf{MKD} &  & \textbf{82.1} \\
        \bottomrule
        \end{tabular}
    }
    \label{tab:compare_kd}
\end{table}

\begin{table}[t]
    \centering
    \caption{\textbf{Distilling ViT-T with teachers of different capacities on ImageNet-1K.} Ordered by model capacity from top to bottom.}
    \resizebox{1.0\linewidth}{!}{
        \begin{tabular}{lcc|c}
        \toprule
        Teacher & \makecell{Params. \\ (M)} & \makecell{Top-1 Acc. \\ (\%)} &  \makecell{Student \\ Top-1 Acc. (\%)} \\
        \midrule
        DeiT-S \cite{deit} & 22.1 & 79.8 & \cellcolor{gray!20}75.9  \\
        CaiT-S24 \cite{cait} & 46.9 & 82.7 & \cellcolor{gray!35}76.4 \\
        DeiT-B-Dist. \cite{deit} & 87.3 & 83.4 & \cellcolor{gray!50}76.8 \\
        BEiT-L \cite{beit} & 304.4 & 87.5 & \cellcolor{gray!65}77.1 \\
        \bottomrule
        \end{tabular}
    }
    \label{tab:fix_student}
\end{table}

\paragraph{Better teachers lead to better students.} In Table \ref{tab:fix_student}, we experiment on distilling ViT-T with teachers of different capacities. We observe that the performance of the student improves as the teachers' capacities increase. Specifically, distilling from BEiT-L, the performance of the student improves 1.2\%, compared with distilling from DeiT-S. 
We achieve 77.1\% with ViT-T by training for 300 epochs, 0.5\% better than previous best result \cite{deit}, which trains for 1000 epoch. 

\paragraph{MKD overcomes degradations.} Prior arts point out that KD suffers from the teacher-student gap and is incompatible with strong data augmentation. Those problems prevent KD from learning from better teachers or advanced data augmentations. In contrast, our MKD mitigates those degradation problems by adaptively adjusting the temperatures. We do not notice the degradation of students even when the teacher is 50$\times$ larger than the student. Note that both the teachers and the students in Table \ref{tab:fix_student} are trained with strong data augmentations, such as Mixup \cite{zhang2017mixup} and CutMix \cite{yun2019cutmix}. Those results demonstrate the robustness and the effectiveness of our MKD.

\begin{table}[t]
    \centering
    \caption{\textbf{Comparison with previous results of ViT on ImageNet-1K.} The previous best results are underlined. }
    \resizebox{1.0\linewidth}{!}{
        \begin{tabular}{l|cccc}
        \toprule
        Method & ViT-T & ViT-S & ViT-B & ViT-L \\
        \midrule
        Train-from-scratch & 72.2 & 79.8 & 81.8 & 82.6 \\
        TrainViT \cite{trainvit} & 73.75 &  80.46 & 83.96 & 83.98 \\
        Manifold \cite{manifold} & 75.1  & 81.5 & - & - \\
        DeiT \cite{deit}         & \underline{76.6}  & \underline{82.6} & \underline{84.2} & - \\
        DINO \cite{dino}     &  & 81.5 & 82.8 & - \\
        MoCo v3 \cite{moco_v3}  & - & - & 83.2 & 84.1 \\
        BEiT \cite{beit}     & - & - & 83.2 & 85.2 \\
        MAE \cite{mae}      & - & - & 83.6 & \underline{85.9} \\
        \midrule
        \textbf{MKD} & \textbf{77.1} & \textbf{83.1} & \textbf{85.1} & \textbf{86.5} \\
        \bottomrule
        \end{tabular}
    }
    \label{tab:sota}
\end{table}

\begin{table}[t]
    \centering
    \caption{\textbf{Comparison with previous distillation methods on CIFAR-100.} Average over 5 runs.}
    \resizebox{1.0\linewidth}{!}{
    \begin{tabular}{lcccc}
        \toprule
        Teacher & ResNet110 & ResNet56 & ResNet110 & ResNet32x4 \\
        Student & ResNet20 & ResNet20 & ResNet32 & ResNet8x4  \\
        \midrule
        Teacher & 77.52 & 76.04 & 77.52 & 81.96 \\
        Student & 68.55 & 68.55 & 70.74 & 71.58  \\
        \midrule
        KD \cite{kd} & 68.71 & 68.89 & 72.34 & 70.15 \\
        CRD \cite{crd} & 71.14 & 71.91 & 73.43 & 73.75 \\
        CRD+KD & 71.06 & 71.98 & 72.54 & 73.55 \\
        \textbf{MKD} & 70.64 & 70.93 & 72.74 & 71.6 \\
        CRD+\textbf{MKD} & \textbf{71.58} & \textbf{72.01} & \textbf{73.75} & \textbf{74.24} \\
        \bottomrule
    \end{tabular}
    }
    \label{tab:cifar100}
\end{table}

\paragraph{Comparisons with previous results of ViT.} In Table \ref{tab:sota}, we compare our distilled ViT models with previous approaches, including distillation, pre-training with large scale datasets followed by finetuning, and self-supervised methods. All MKD results are obtained by distilling BEiT-L (87.5\%).

With MKD, we achieve the best performance on popular ViT architectures among compared methods that use only ImageNet-1K as training data, across tiny to large models. Our MKD distilled ViT-L obtains 86.5\% accuracy, +3.9\% better than training without distillation \cite{mae}. Compared with MAE that employs self-supervised training for 1,600 epochs, our MKD achieves better results with much fewer training epochs.

Training large ViT models is nontrivial \cite{trainvit}, and
previous works proposed different techniques \cite{deit,mae} to achieved good results. We provide a promising approach to train ViTs by distillation, which is more accurate, simpler, and faster.

\subsection{CIFAR-100 Results}

In Table \ref{tab:cifar100}, we compare our MKD with other distillation methods on CIFAR-100 with different teacher-student pairs. Our method outperforms compared methods for different teacher-student pairs.

Besides using MKD alone, we also experiment with combining CRD \cite{crd} and the proposed MKD. As shown in Table \ref{tab:cifar100}, compared to CRD, vanilla KD produces worse results when combined with CRD. In contrast, we obtain consistent gains over the strong baseline.

\section{Ablation Studies}
\label{sec:ablation_study}

In this section, we investigate individual components of our Meta Knowledge Distillation on the CIFAR-100 benchmark. The ablation studies are conducted with the setup of distilling ResNet32x4 to ResNet8x4. The teacher and the student are trained with CutMix for data augmentation. For other details, please see more details in Section \ref{sec:experimental_setup}. 

\paragraph{MKD vs. grid search} We list the baseline results in Table \ref{tab:ablation_baseline}, including the results of vanilla Knowledge Distillation (KD), CRD \cite{crd}, and our MKD. 
We see that the default KD setting provided by CRD fail to improve the student (71.2\% vs. the train-from-scratch result of 73.4\%). However, after grid searching the temperatures, vanilla KD obtains 73.7\%, being slightly better than the train-from-scratch result. 
In contrast, our MKD improves our grid searched results, and achieves 74.4\%.

\begin{figure}[t]
    \centering
    \includegraphics[width=0.8\linewidth]{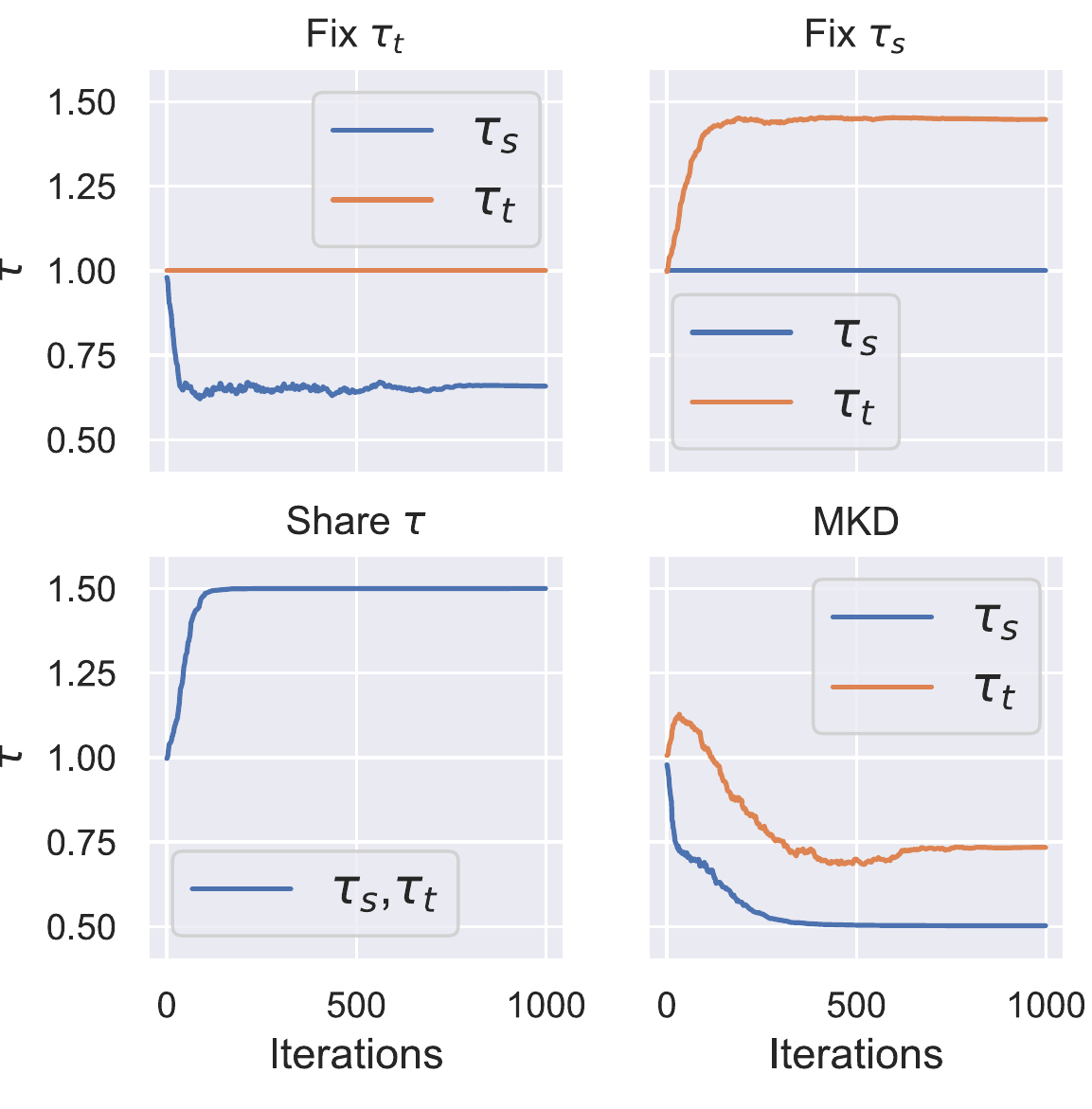}
    \vspace{-1em}
    \caption{The evolution curves of temperatures during training.}
    \label{fig:tau_change}
\end{figure}

\begin{table}[t]
    \centering
    \caption{Baseline results of ablation studies.}
    \resizebox{0.9\linewidth}{!}{
    \begin{tabular}{l|c}
    \toprule
    Method & Top-1 Acc. (\%) \\
    \midrule
    Teacher (ResNet32x4) & 81.9 \\
    Student (ResNet8x4) & 73.4 \\
    \midrule
    KD (CRD) \cite{kd} & 71.2 \\
    CRD \cite{crd} & 70.9 \\
    KD (grid search) \cite{kd} & 73.7 \\
    \textbf{MKD} & \textbf{74.4} \\
    
    \bottomrule
    \end{tabular}
    }
    \label{tab:ablation_baseline}
\end{table}

\begin{table}[t]
    \centering
    \begin{minipage}[t]{0.48\linewidth}
        \caption{Meta-learning targets. Learning different temperatures for the teacher and the student achieves better results. }
        \centering
        \resizebox{1.0\linewidth}{!}{
        \begin{tabular}{l|c}
        \toprule
        Setup & \makecell{Top-1 Acc. \\ (\%)} \\
        \midrule
        Fix $\tau_t$, learn $\tau_s$ & 74.0 \\
        Fix $\tau_s$, learn $\tau_t$ & 72.9 \\
        Share learnable $\tau$ & 72.6 \\
        Learn $\{\tau_s,\tau_t\}$ & \textbf{74.4} \\
        \bottomrule
        \end{tabular}
        }
        \label{tab:ablation_learn_temp}
    \end{minipage}
    \begin{minipage}[t]{0.48\linewidth}
        \caption{Initial temperature. $\Delta$ represents accuracy difference after applying MKD.}
        \centering
        \resizebox{1.0\linewidth}{!}{
        \begin{tabular}{c|cc}
        \toprule
        Benchmark & $\tau_{init}$ & \makecell{$\Delta$ acc. (\%)} \\
        \midrule
        \multirow{4}{*}{CIFAR-100} & 1 & 0.7 \\
        & 2 & 0.7 \\
        & 3 & 0.5 \\
        & 4 & 0.6 \\ 
        \midrule
        \multirow{4}{*}{ImageNet-1K} & 1 & 0.5 \\ 
        & 2 & 2.4 \\ 
        & 3 & 2.6 \\ 
        & 4 & 2.1 \\ 
        \bottomrule
        \end{tabular}
        }
        \label{tab:ablation_init_temp}
    \end{minipage}
\end{table}

\begin{table}[t]
    \centering
    \begin{minipage}[t]{0.48\linewidth}
        \caption{Design of temperature output approaches.}
        \centering
        \resizebox{1.0\linewidth}{!}{
        \begin{tabular}{l|c}
        \toprule
        Setup & Top-1 Acc. (\%) \\
        \midrule
        Learnable parameter & 73.4 \\
        MLP network & \textbf{74.4} \\
        \bottomrule
        \end{tabular}
        }
        \label{tab:ablation_tau}
    \end{minipage}
    \begin{minipage}[t]{0.48\linewidth}
        \caption{Design of meta loss function.}
        \centering
        \resizebox{1.0\linewidth}{!}{
        \begin{tabular}{c|cc}
        \toprule
        Meta loss function & Top-1 Acc. (\%) \\
        \midrule
        cross-entropy loss & 74 \\ 
        Equation (\ref{eq:meta_loss}) & \textbf{74.4} \\
        \bottomrule
        \end{tabular}
        }
        \label{tab:ablation_meta_objective}
    \end{minipage}
\end{table}

\begin{table}[t]
    \centering
    \caption{Applying grid searched temperature of source dataset to target dataset.}
    \resizebox{0.75\linewidth}{!}{
        \begin{tabular}{cc|c}
        \toprule
        Source & Target & Top-1 Acc. (\%) \\
        \midrule
        CIFAR-100 & ImageNet-1K & 70.1 \\
        ImageNet-1K & ImageNet-1K & \textbf{75.9} \\
        \midrule
        ImageNet-1K & CIFAR-100 & 71.4 \\
        CIFAR-100 & CIFAR-100 & \textbf{73.7} \\
        \bottomrule
        \end{tabular}
    }
    \label{tab:transfer}
\end{table}

\paragraph{Meta-learning targets.} The meta-learning targets of MKD can be flexibly designed, as shown in Table \ref{tab:ablation_learn_temp}. As for the capacity gap between the teacher and the student, using different temperatures for them allows more flexible control of the their softness of outputs. We see that jointly learning separate temperatures for the teacher and the student achieves best results. In contrast, using a shared temperatures for both the teacher and student is widely used in previous works and leads to inferior performances,
which demonstrates that a shared temperature is insufficient for better knowledge transfer. We further visualize the adaptation of the temperatures over the course of training in Figure \ref{fig:tau_change}.

\paragraph{Initial temperatures.} The influence of the initial temperature is studied in Table \ref{tab:ablation_init_temp}. We experiment with the widely used temperatures, and presents the accuracy difference after applying MKD. We see that MKD is robust to initial temperatures, and obtains consistent performance gains over the initial temperatures. Note that for ImageNet-1K, comparing to the best grid-searched temperature, MKD also obtains 0.5\% accuracy gain.

\paragraph{Ways to generate temperatures.} We experiment on two ways to generate temperatures, 1) two learnable parameters directly updated via back-propagation, and 2) generating temperatures from a small neural network, as expressed in Equation \eqref{eq:generate_tau}. The later approach obtains 1\% performance gain compared with the former.

Previous meta-hyperparameter-optimization works \cite{xu2018meta, metasgd, baik2020meta,khodak2019adaptive} usually employ the first approach to optimize their hyperparameters. In contrast, we propose to generate the hyperparameters to be learned with a small neural network, e.g. 2-layer MLP network. The neural network has larger capacity, and is able to adapt faster, leading to better performance. Baik et al. \cite{baik2020meta} employ similar strategy to generate hyperparameters, but requires additional state as the input.

\paragraph{Meta loss functions.} We compare different meta loss functions in Table \ref{tab:ablation_meta_objective}. Using cross-entropy (CE) loss on the validation set to optimize is a natural choice, and has been utilized in previous works \cite{mpl}. However, as stated in Section \ref{sec:meta_objective}, the CE loss may not indicate the true performance. As a result, we propose an alternative that only optimizes the incorrect samples as express in Equation \eqref{eq:meta_loss}, which leads to better results on CIFAR-100.

\paragraph{Transfer temperatures across dataset}
We apply the grid searched temperatures on a source dataset and transfer the searched temperatures to to a target dataset. As shown in Table \ref{tab:transfer}, transferring the temperatures leads to inferior performance. For instance, directly searching on ImageNet-1K is 5.8\% better than transferring from CIFAR-100. The results also suggest that the temperatures should be adjusted for different datasets.
The results provide the strong justification of adaptively and separately searching the temperatures for different distillation setups.

\section{Related Works}
\label{sec:related_works}
\paragraph{Knowledge distillation}
The initial idea of knowledge distillation (KD) is introduced by the work of \cite{modlecompression} and \cite{kd}, attempting to transfer the knowledge of a cumbersome model (teacher) to a lightweight model (student). Buciluǎ et al. \yrcite{modlecompression} achieve the knowledge transferring by matching logits. Hinton et al. \yrcite{kd} propose to soft the logits before softmax with a temperature, and match the soften probabilities of the teacher and the student for distillation. The soften probabilities provide useful information about the learned representation of the teacher model, and the temperature is able to control which part of the information to focus on. After that, a lot of works try to improve KD with various techniques. FitNets \cite{fitnet} proposes to match the intermediate features for better knowledge transfer. Zagoruyko et al. \cite{zagoruyko2016paying} transfer the attention map of intermediate features from the teacher to the student. CRD \cite{crd} introduce contrastive objective for representation transfer, which achieves state-of-the-art results. Other papers \cite{park2019relational, yim2017gift, huang2017like, kim2018paraphrasing, ahn2019variational, koratana2019lit} have proposed various distillation criteria based on representations. 

\paragraph{Meta-learning for hyperparameter optimization}
Hyperparameters are hard to tune manually, and lots of works try to automate the tuning by meta-learning. Meta-SGD \cite{metasgd} optimize the learning rate along model parameters. Xu et al. \cite{xu2018meta} propose to learn the return function in Reinforcement Learning with tunable meta-parameters, and use specialization form for different scenarios.
Baik et al. \cite{baik2020meta} propose to update the hyperparameters (i.e. learning rate and weight decay coefficients) with meta-optimization. Their work uses the learning state as the input of the meta-model to produce the hyperparameters, but our work does not.

\section{Conclusion}
\label{sec:conclusion}

In this paper, we propose Meta Knowledge Distillation to meta-learn the distillation loss with learnable temperatures. While recent works point out that knowledge distillation suffers from two degradation problems, our study show that those problems can be much mitigated with proper temperatures to the teacher and the student. Our MKD can adaptively adjust the temperatures over the training process according to the gradients of the
learning objective. With extensive experiments, we show that our MKD is robust to different scales of the datasets, architectures of the teacher or student models, and different types of data augmentation. We achieve the best performance with popular ViT architectures among compared methods that use only ImageNet-1K as training data, across tiny to large models. 


\bibliography{example_paper}

\begin{thebibliography}{43}
\providecommand{\natexlab}[1]{#1}
\providecommand{\url}[1]{\texttt{#1}}
\expandafter\ifx\csname urlstyle\endcsname\relax
  \providecommand{\doi}[1]{doi: #1}\else
  \providecommand{\doi}{doi: \begingroup \urlstyle{rm}\Url}\fi

\bibitem[Ahn et~al.(2019)Ahn, Hu, Damianou, Lawrence, and
  Dai]{ahn2019variational}
Ahn, S., Hu, S.~X., Damianou, A., Lawrence, N.~D., and Dai, Z.
\newblock Variational information distillation for knowledge transfer.
\newblock In \emph{Proceedings of the IEEE/CVF Conference on Computer Vision
  and Pattern Recognition}, pp.\  9163--9171, 2019.

\bibitem[Ba et~al.(2016)Ba, Kiros, and Hinton]{ba2016layer}
Ba, J.~L., Kiros, J.~R., and Hinton, G.~E.
\newblock Layer normalization.
\newblock \emph{arXiv preprint arXiv:1607.06450}, 2016.

\bibitem[Baik et~al.(2020)Baik, Choi, Choi, Kim, and Lee]{baik2020meta}
Baik, S., Choi, M., Choi, J., Kim, H., and Lee, K.~M.
\newblock Meta-learning with adaptive hyperparameters.
\newblock \emph{arXiv preprint arXiv:2011.00209}, 2020.

\bibitem[Bao et~al.(2021)Bao, Dong, and Wei]{beit}
Bao, H., Dong, L., and Wei, F.
\newblock Beit: Bert pre-training of image transformers.
\newblock \emph{arXiv preprint arXiv:2106.08254}, 2021.

\bibitem[Beyer et~al.(2021)Beyer, Zhai, Royer, Markeeva, Anil, and
  Kolesnikov]{beyer2021knowledge}
Beyer, L., Zhai, X., Royer, A., Markeeva, L., Anil, R., and Kolesnikov, A.
\newblock Knowledge distillation: A good teacher is patient and consistent.
\newblock \emph{arXiv preprint arXiv:2106.05237}, 2021.

\bibitem[Buciluǎ et~al.(2006)Buciluǎ, Caruana, and
  Niculescu-Mizil]{modlecompression}
Buciluǎ, C., Caruana, R., and Niculescu-Mizil, A.
\newblock Model compression.
\newblock In \emph{Proceedings of the 12th ACM SIGKDD international conference
  on Knowledge discovery and data mining}, pp.\  535--541, 2006.

\bibitem[Caron et~al.(2021)Caron, Touvron, Misra, J{\'e}gou, Mairal,
  Bojanowski, and Joulin]{dino}
Caron, M., Touvron, H., Misra, I., J{\'e}gou, H., Mairal, J., Bojanowski, P.,
  and Joulin, A.
\newblock Emerging properties in self-supervised vision transformers.
\newblock \emph{arXiv preprint arXiv:2104.14294}, 2021.

\bibitem[Chen et~al.(2021)Chen, Xie, and He]{moco_v3}
Chen, X., Xie, S., and He, K.
\newblock An empirical study of training self-supervised vision transformers.
\newblock \emph{arXiv preprint arXiv:2104.02057}, 2021.

\bibitem[Cho \& Hariharan(2019)Cho and Hariharan]{earlystopkd}
Cho, J.~H. and Hariharan, B.
\newblock On the efficacy of knowledge distillation.
\newblock In \emph{Proceedings of the IEEE/CVF International Conference on
  Computer Vision}, pp.\  4794--4802, 2019.

\bibitem[Cubuk et~al.(2020)Cubuk, Zoph, Shlens, and Le]{randaug}
Cubuk, E.~D., Zoph, B., Shlens, J., and Le, Q.~V.
\newblock Randaugment: Practical automated data augmentation with a reduced
  search space.
\newblock In \emph{Proceedings of the IEEE/CVF Conference on Computer Vision
  and Pattern Recognition Workshops}, pp.\  702--703, 2020.

\bibitem[Cui \& Yan(2021)Cui and Yan]{cui2021isotonic}
Cui, W. and Yan, S.
\newblock Isotonic data augmentation for knowledge distillation.
\newblock \emph{arXiv preprint arXiv:2107.01412}, 2021.

\bibitem[Das et~al.(2020)Das, Massa, Kulkarni, and
  Rekatsinas]{das2020empirical}
Das, D., Massa, H., Kulkarni, A., and Rekatsinas, T.
\newblock An empirical analysis of the impact of data augmentation on knowledge
  distillation.
\newblock \emph{arXiv preprint arXiv:2006.03810}, 2020.

\bibitem[Deng et~al.(2009)Deng, Dong, Socher, Li, Li, and Fei-Fei]{imagenet}
Deng, J., Dong, W., Socher, R., Li, L.-J., Li, K., and Fei-Fei, L.
\newblock Imagenet: A large-scale hierarchical image database.
\newblock In \emph{2009 IEEE conference on computer vision and pattern
  recognition}, pp.\  248--255. Ieee, 2009.

\bibitem[Dosovitskiy et~al.(2020)Dosovitskiy, Beyer, Kolesnikov, Weissenborn,
  Zhai, Unterthiner, Dehghani, Minderer, Heigold, Gelly, et~al.]{vit}
Dosovitskiy, A., Beyer, L., Kolesnikov, A., Weissenborn, D., Zhai, X.,
  Unterthiner, T., Dehghani, M., Minderer, M., Heigold, G., Gelly, S., et~al.
\newblock An image is worth 16x16 words: Transformers for image recognition at
  scale.
\newblock \emph{arXiv preprint arXiv:2010.11929}, 2020.

\bibitem[He et~al.(2016)He, Zhang, Ren, and Sun]{resnet}
He, K., Zhang, X., Ren, S., and Sun, J.
\newblock Deep residual learning for image recognition.
\newblock In \emph{Proceedings of the IEEE conference on computer vision and
  pattern recognition}, pp.\  770--778, 2016.

\bibitem[He et~al.(2021)He, Chen, Xie, Li, Doll{\'a}r, and Girshick]{mae}
He, K., Chen, X., Xie, S., Li, Y., Doll{\'a}r, P., and Girshick, R.
\newblock Masked autoencoders are scalable vision learners.
\newblock \emph{arXiv preprint arXiv:2111.06377}, 2021.

\bibitem[Hinton et~al.(2015)Hinton, Vinyals, and Dean]{kd}
Hinton, G., Vinyals, O., and Dean, J.
\newblock Distilling the knowledge in a neural network.
\newblock \emph{arXiv preprint arXiv:1503.02531}, 2015.

\bibitem[Huang et~al.(2016)Huang, Sun, Liu, Sedra, and Weinberger]{droppath}
Huang, G., Sun, Y., Liu, Z., Sedra, D., and Weinberger, K.~Q.
\newblock Deep networks with stochastic depth.
\newblock In \emph{European conference on computer vision}, pp.\  646--661.
  Springer, 2016.

\bibitem[Huang \& Wang(2017)Huang and Wang]{huang2017like}
Huang, Z. and Wang, N.
\newblock Like what you like: Knowledge distill via neuron selectivity
  transfer.
\newblock \emph{arXiv preprint arXiv:1707.01219}, 2017.

\bibitem[Jia et~al.(2021)Jia, Han, Wang, Tang, Guo, Zhang, and Tao]{manifold}
Jia, D., Han, K., Wang, Y., Tang, Y., Guo, J., Zhang, C., and Tao, D.
\newblock Efficient vision transformers via fine-grained manifold distillation.
\newblock \emph{arXiv preprint arXiv:2107.01378}, 2021.

\bibitem[Khodak et~al.(2019)Khodak, Balcan, and Talwalkar]{khodak2019adaptive}
Khodak, M., Balcan, M.-F., and Talwalkar, A.
\newblock Adaptive gradient-based meta-learning methods.
\newblock \emph{arXiv preprint arXiv:1906.02717}, 2019.

\bibitem[Kim et~al.(2018)Kim, Park, and Kwak]{kim2018paraphrasing}
Kim, J., Park, S., and Kwak, N.
\newblock Paraphrasing complex network: Network compression via factor
  transfer.
\newblock \emph{arXiv preprint arXiv:1802.04977}, 2018.

\bibitem[Koratana et~al.(2019)Koratana, Kang, Bailis, and
  Zaharia]{koratana2019lit}
Koratana, A., Kang, D., Bailis, P., and Zaharia, M.
\newblock Lit: Learned intermediate representation training for model
  compression.
\newblock In \emph{International Conference on Machine Learning}, pp.\
  3509--3518. PMLR, 2019.

\bibitem[Krizhevsky et~al.(2009)Krizhevsky, Hinton, et~al.]{cifar}
Krizhevsky, A., Hinton, G., et~al.
\newblock Learning multiple layers of features from tiny images.
\newblock 2009.

\bibitem[Li et~al.(2017)Li, Zhou, Chen, and Li]{metasgd}
Li, Z., Zhou, F., Chen, F., and Li, H.
\newblock Meta-sgd: Learning to learn quickly for few-shot learning.
\newblock \emph{arXiv preprint arXiv:1707.09835}, 2017.

\bibitem[Loshchilov \& Hutter(2017)Loshchilov and Hutter]{adamw}
Loshchilov, I. and Hutter, F.
\newblock Decoupled weight decay regularization.
\newblock \emph{arXiv preprint arXiv:1711.05101}, 2017.

\bibitem[Mirzadeh et~al.(2020)Mirzadeh, Farajtabar, Li, Levine, Matsukawa, and
  Ghasemzadeh]{takd}
Mirzadeh, S.~I., Farajtabar, M., Li, A., Levine, N., Matsukawa, A., and
  Ghasemzadeh, H.
\newblock Improved knowledge distillation via teacher assistant.
\newblock In \emph{Proceedings of the AAAI Conference on Artificial
  Intelligence}, volume~34, pp.\  5191--5198, 2020.

\bibitem[M{\"u}ller et~al.(2019)M{\"u}ller, Kornblith, and
  Hinton]{muller2019does}
M{\"u}ller, R., Kornblith, S., and Hinton, G.
\newblock When does label smoothing help?
\newblock \emph{arXiv preprint arXiv:1906.02629}, 2019.

\bibitem[Park et~al.(2019)Park, Kim, Lu, and Cho]{park2019relational}
Park, W., Kim, D., Lu, Y., and Cho, M.
\newblock Relational knowledge distillation.
\newblock In \emph{Proceedings of the IEEE/CVF Conference on Computer Vision
  and Pattern Recognition}, pp.\  3967--3976, 2019.

\bibitem[Pham et~al.(2021)Pham, Dai, Xie, and Le]{mpl}
Pham, H., Dai, Z., Xie, Q., and Le, Q.~V.
\newblock Meta pseudo labels.
\newblock In \emph{Proceedings of the IEEE/CVF Conference on Computer Vision
  and Pattern Recognition}, pp.\  11557--11568, 2021.

\bibitem[Romero et~al.(2014)Romero, Ballas, Kahou, Chassang, Gatta, and
  Bengio]{fitnet}
Romero, A., Ballas, N., Kahou, S.~E., Chassang, A., Gatta, C., and Bengio, Y.
\newblock Fitnets: Hints for thin deep nets.
\newblock \emph{arXiv preprint arXiv:1412.6550}, 2014.

\bibitem[Shen et~al.(2021)Shen, Liu, Xu, Chen, Cheng, and
  Savvides]{shen2021label}
Shen, Z., Liu, Z., Xu, D., Chen, Z., Cheng, K.-T., and Savvides, M.
\newblock Is label smoothing truly incompatible with knowledge distillation: An
  empirical study.
\newblock \emph{arXiv preprint arXiv:2104.00676}, 2021.

\bibitem[Steiner et~al.(2021)Steiner, Kolesnikov, Zhai, Wightman, Uszkoreit,
  and Beyer]{trainvit}
Steiner, A., Kolesnikov, A., Zhai, X., Wightman, R., Uszkoreit, J., and Beyer,
  L.
\newblock How to train your vit? data, augmentation, and regularization in
  vision transformers.
\newblock \emph{arXiv preprint arXiv:2106.10270}, 2021.

\bibitem[Szegedy et~al.(2016)Szegedy, Vanhoucke, Ioffe, Shlens, and
  Wojna]{inception}
Szegedy, C., Vanhoucke, V., Ioffe, S., Shlens, J., and Wojna, Z.
\newblock Rethinking the inception architecture for computer vision.
\newblock In \emph{Proceedings of the IEEE conference on computer vision and
  pattern recognition}, pp.\  2818--2826, 2016.

\bibitem[Tian et~al.(2019)Tian, Krishnan, and Isola]{crd}
Tian, Y., Krishnan, D., and Isola, P.
\newblock Contrastive representation distillation.
\newblock \emph{arXiv preprint arXiv:1910.10699}, 2019.

\bibitem[Touvron et~al.(2021{\natexlab{a}})Touvron, Cord, Douze, Massa,
  Sablayrolles, and J{\'e}gou]{deit}
Touvron, H., Cord, M., Douze, M., Massa, F., Sablayrolles, A., and J{\'e}gou,
  H.
\newblock Training data-efficient image transformers \& distillation through
  attention.
\newblock In \emph{International Conference on Machine Learning}, pp.\
  10347--10357. PMLR, 2021{\natexlab{a}}.

\bibitem[Touvron et~al.(2021{\natexlab{b}})Touvron, Cord, Sablayrolles,
  Synnaeve, and J{\'e}gou]{cait}
Touvron, H., Cord, M., Sablayrolles, A., Synnaeve, G., and J{\'e}gou, H.
\newblock Going deeper with image transformers.
\newblock \emph{arXiv preprint arXiv:2103.17239}, 2021{\natexlab{b}}.

\bibitem[Vaswani et~al.(2017)Vaswani, Shazeer, Parmar, Uszkoreit, Jones, Gomez,
  Kaiser, and Polosukhin]{vaswani2017attention}
Vaswani, A., Shazeer, N., Parmar, N., Uszkoreit, J., Jones, L., Gomez, A.~N.,
  Kaiser, {\L}., and Polosukhin, I.
\newblock Attention is all you need.
\newblock In \emph{Advances in neural information processing systems}, pp.\
  5998--6008, 2017.

\bibitem[Xu et~al.(2018)Xu, van Hasselt, and Silver]{xu2018meta}
Xu, Z., van Hasselt, H., and Silver, D.
\newblock Meta-gradient reinforcement learning.
\newblock \emph{arXiv preprint arXiv:1805.09801}, 2018.

\bibitem[Yim et~al.(2017)Yim, Joo, Bae, and Kim]{yim2017gift}
Yim, J., Joo, D., Bae, J., and Kim, J.
\newblock A gift from knowledge distillation: Fast optimization, network
  minimization and transfer learning.
\newblock In \emph{Proceedings of the IEEE Conference on Computer Vision and
  Pattern Recognition}, pp.\  4133--4141, 2017.

\bibitem[Yun et~al.(2019)Yun, Han, Oh, Chun, Choe, and Yoo]{yun2019cutmix}
Yun, S., Han, D., Oh, S.~J., Chun, S., Choe, J., and Yoo, Y.
\newblock Cutmix: Regularization strategy to train strong classifiers with
  localizable features.
\newblock In \emph{Proceedings of the IEEE/CVF International Conference on
  Computer Vision}, pp.\  6023--6032, 2019.

\bibitem[Zagoruyko \& Komodakis(2016)Zagoruyko and
  Komodakis]{zagoruyko2016paying}
Zagoruyko, S. and Komodakis, N.
\newblock Paying more attention to attention: Improving the performance of
  convolutional neural networks via attention transfer.
\newblock \emph{arXiv preprint arXiv:1612.03928}, 2016.

\bibitem[Zhang et~al.(2017)Zhang, Cisse, Dauphin, and
  Lopez-Paz]{zhang2017mixup}
Zhang, H., Cisse, M., Dauphin, Y.~N., and Lopez-Paz, D.
\newblock mixup: Beyond empirical risk minimization.
\newblock \emph{arXiv preprint arXiv:1710.09412}, 2017.

\end{thebibliography}
\bibliographystyle{icml2022}

\newpage
\appendix
\onecolumn
\section{Appendix}
\label{sec:appendix}

\subsection{Training details on CIFAR-100}
The train setting on CIFAR-100 is in Table \ref{tab:cifar_traing_setting}.

\begin{table}[h]
    \centering
    \caption{Training settings on CIFAR-100. $\ast$ optional config.}
    \begin{tabular}{l|c}
        \toprule
        config & value \\
        \midrule
        optimizer & SGD \\
        learning rate & 0.05 \\
        weight decay & 0.0005 \\
        batch size & 64 \\
        learning rate schedule & cosine decay \\
        training epochs & 300 \\
        augmentation & random crop and flip \\
        LabelSmooth \textsuperscript{$\ast$} \cite{inception} & 0.1 \\
        Mixup \textsuperscript{$\ast$} \cite{zhang2017mixup} & 0.2 \\
        CutMix \textsuperscript{$\ast$} \cite{yun2019cutmix} & 1.0 \\
        \bottomrule
    \end{tabular}
    \label{tab:cifar_traing_setting}
\end{table}

\subsection{Training details on ImageNet}
We follow the standard ViT architecture \cite{vit}. It consists of a stack of Transformer blocks \cite{vaswani2017attention}, and each block consists of a multi-head self-attention block and an FFN block, with LayerNorm\cite{ba2016layer}. The ViT-T and ViT-S are introduced in DeiT \cite{deit}. 

Our training of ViT follows common practice of supervised ViT training. The default setting is in Table \ref{tab:traing_setting}.

\begin{table}[h]
    \centering
    \caption{Training settings on ImageNet-1K.}
    \begin{tabular}{l|c}
        \toprule
        config & value \\
        \midrule
        optimizer & AdamW \cite{adamw} \\
        learning rate & 0.001 (T/S/B), 0.0003 (L) \\
        weight decay & 0.05 \\
        batch size & 1024 \\
        learning rate schedule & cosine decay \\
        warmup epochs & 5 \\
        training epochs & 300 (T/S), 600 (S/B/L) \\
        augmentation & RandAug(9, 0.5) \cite{randaug} \\
        LabelSmooth \cite{inception} & 0.1 \\
        Mixup \cite{zhang2017mixup} & 0.8 \\
        CutMix \cite{yun2019cutmix} & 1.0 \\
        drop path \cite{droppath} & 0.0 (T/S/B), 0.3 (L) \\
        \bottomrule
    \end{tabular}
    \label{tab:traing_setting}
\end{table}

\subsection{Training details of meta-parameters}

For our proposed temperature prediction network, we employ a 2-layer MLP with ReLU activation between the layers. The output is scaled with sigmoid function. The input embedding dimension is set to 8, and the hidden dimension is set to 16.

The default setting to train meta-parameters is in Table \ref{tab:traing_setting}.

\begin{table}[h]
    \centering
    \caption{Training settings of meta-parameters}
    \begin{tabular}{l|c}
        \toprule
        config & value \\
        \midrule
        optimizer & AdamW \cite{adamw} \\
        learning rate & 0.0003 \\
        weight decay & 0.00005 \\
        batch size & 1024 (ImageNet), 64 (CIFAR)  \\
        learning rate schedule & cosine decay \\
        \bottomrule
    \end{tabular}
    \label{tab:my_label}
\end{table}


\end{document}